\documentclass[sigconf]{acmart}

\usepackage{xcolor}
\usepackage{multirow}
\usepackage{booktabs}
\usepackage{pifont}

\ifdefined\draftversion
    \newcommand{\pol}[1]{\textcolor{blue}{(Pol Comment: #1)}}
    \newcommand{\eri}[1]{\textcolor{magenta}{(Eri: #1)}}
    
    \newcommand{\ab}[1]{\textcolor{green}{(Abhijit Comment: #1)}}
    \newcommand{\axl}[1]{\textcolor{cyan}{(Axel Comment: #1)}}
    \newcommand{\mau}[1]{\textcolor{purple}{(Mau: #1)}}
\else
    \newcommand{\pol}[1]{}
    \newcommand{\eri}[1]{}
    
    \newcommand{\ab}[1]{}
    \newcommand{\axl}[1]{}
    \newcommand{\mau}[1]{}
\fi

\usepackage{amsmath}

\DeclareMathOperator*{\argmin}{arg\,min}

\newcommand{\ie}{\textit{i.e.} }

\usepackage[ruled]{algorithm2e}
\usepackage{array}

\begin{document}
\title{A Data-Driven Approach to Dataflow-Aware Online~Scheduling~for~Graph~Neural~Network~Inference}
\author{Pol Puigdemont}
\affiliation{
\institution{Universitat Politècnica de Catalunya}
\country{Spain}
}
\authornote{Equal Contribution (e-mails: polpuigdemont@gmail.com, enrico.russo@phd.unict.it)}

\author{Enrico Russo}
\affiliation{
\institution{University of Catania}
\country{Italy}
}
\authornotemark[1]

\author{Axel Wassington}
\affiliation{
\institution{Universitat Politècnica de Catalunya}
\country{Spain}
}

\author{Abhijit Das}
\affiliation{
\institution{Universitat Politècnica de Catalunya}
\country{Spain}
}

\author{Sergi Abadal}
\affiliation{
\institution{Universitat Politècnica de Catalunya}
\country{Spain}
}

\author{Maurizio Palesi}
\affiliation{
\institution{University of Catania}
\country{Italy}
}
\renewcommand{\shortauthors}{Puigdemont and Russo, et al.}
\begin{abstract}
  Graph Neural Networks (GNNs) have shown significant promise in various domains, such as recommendation systems, bioinformatics, and network analysis. However, the irregularity of graph data poses unique challenges for efficient computation, leading to the development of specialized GNN accelerator architectures that surpass traditional CPU and GPU performance. Despite this, the structural diversity of input graphs results in varying performance across different GNN accelerators, depending on their dataflows. This variability in performance due to differing dataflows and graph properties remains largely unexplored, limiting the adaptability of GNN accelerators.
  To address this, we propose a data-driven framework for dataflow-aware latency prediction in GNN inference. Our approach involves training regressors to predict the latency of executing specific graphs on particular dataflows, using simulations on synthetic graphs. Experimental results indicate that our regressors can predict the optimal dataflow for a given graph with up to 91.28\% accuracy and a Mean Absolute Percentage Error (MAPE) of 3.78\%. Additionally, we introduce an online scheduling algorithm that uses these regressors to enhance scheduling decisions. Our experiments demonstrate that this algorithm achieves up to $3.17\times$ speedup in mean completion time and $6.26\times$ speedup in mean execution time compared to the best feasible baseline across all datasets.
\end{abstract}


\settopmatter{printacmref=false}
\setcopyright{none}
\renewcommand\footnotetextcopyrightpermission[1]{}
\pagestyle{plain}

\maketitle

\section{Introduction}
GNNs have emerged as powerful tools for tackling a diverse range of tasks in fields of a relational nature, such as recommendation systems, bioinformatics, network analysis and quantum computing. Graph data is inherently irregular when compared to other sources, such as images or text. This irregularity introduces unique challenges related to the data movement and memory accesses involved in computing GNNs. These unique challenges have motivated the rise of specific GNN accelerator architectures which outperform conventional CPU and GPU acceleration, allowing for a much faster computation enabling low-latency GNN applications \cite{abadal_acmcs22}.

\begin{figure}[t]
    \centering
    \includegraphics[width=0.9\linewidth,trim={2mm 2mm 0 0},clip]{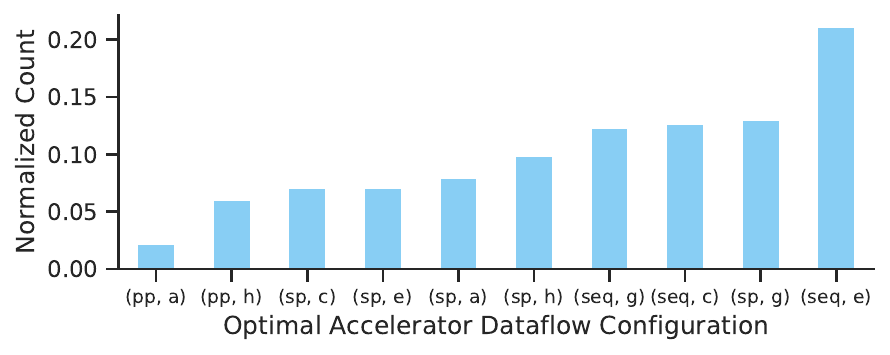}
    \vspace{-8pt}
    \caption{Latency-optimal GNN accelerator dataflow configuration distribution (out of 24) for the ENZYMES dataset \cite{tudataset}.}
    \label{fig:motivation}
    \vspace{-0.6cm}
\end{figure}

Despite the advancements in accelerator design for GNNs, the structural diversity of the input graphs still poses a substantial challenge. Variations in the graph's structure can lead to significant performance fluctuations across different dataflow strategies employed by accelerators. In Figure~\ref{fig:motivation}, we show the distribution of optimal dataflows in terms of latency for an example dataset among the considered possible configurations that will be introduced in the rest of this work. It can be seen that even for graphs belonging to the same domain, the selection of the best configuration is not trivial, and there is no one-size-fits-all solution.

In response to this issue, we propose a novel framework for data-driven dataflow-aware latency prediction for GNN inference. Our framework leverages lightweight parametric regressors trained on a large dataset of synthetic graphs to predict the latency of GNN inference under various dataflow strategies. This predictive capability allows for the optimization of GNN execution by selecting the most suitable dataflow strategy for a given graph structure. Additionally, we introduce an online scheduling algorithm that utilizes our latency prediction framework to dynamically allocate GNN inference tasks across heterogeneous accelerators. Our contribution can be summarized in the following points:

\begin{itemize}
    \item A new dataset generation method for GNN inference computation on different accelerator dataflows.
    \item A novel model for efficient latency estimation.
    \item A predictor-informed online scheduling algorithm for heterogeneous multi-accelerator settings.
\end{itemize}

We conduct extensive experiments to evaluate the effectiveness of our proposed solutions. On the one hand, we test the predictors across multiple axes by measuring the regression accuracy, ranking accuracy and the latency of a single configurable accelerator computing the GNN with the best dataflow according to the model. Experiments show that our model achieves up to $91.28\%$ accuracy and $3.78$ regression MAPE in predicting the best configuration. This leads to a notable single accelerator improvement of up to $93.63\%$, a $58.42\%$ improvement over the best fixed configuration for the dataset and $0.56\%$ degradation over optimal on both real graph and synthetic datasets. On the other hand, we evaluate the performance of our online scheduling algorithm within a heterogeneous multi-accelerator environment where GNNs that arrive in a queue for computation are assigned to an accelerator. The experimental results reveal significant gains in scheduling performance with a $3.17\times$ speedup in mean completion time, $6.26\times$ speedup in mean execution time and more than $1000\times$ speedup in turnaround time against the best feasible baseline on both synthetic and real-world datasets.

\section{Background and Motivation}

\subsection{Graph Neural Networks}
\begin{figure}[t]
    \centering
    \includegraphics[width=0.9\linewidth]{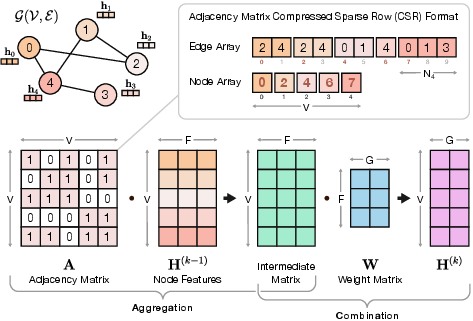}
    \vspace{-9pt}
    \caption{Matrix representation of the GNN $k$-th layer aggregation and combination phases \cite{garg2022understanding}.}
    \label{fig:gnn}
    \vspace{-0.5cm}
\end{figure}

\begin{figure*}[t]
    \centering
    \includegraphics[width=\textwidth]{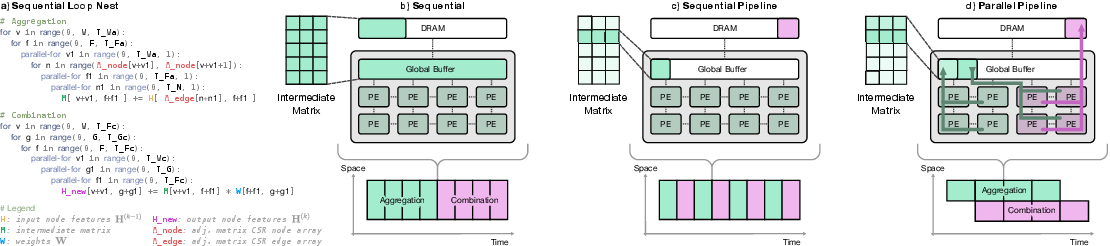}
    \vspace{-15pt}
    \caption{Loop nest representation and inter-phase dataflows with corresponding memory usages.}
    \label{fig:dataflow}
\end{figure*}

GNNs are a family of algorithms designed to apply machine learning techniques to relational data, falling under the broader category of geometric deep learning algorithms. The GNN model is structured in layers, where each layer's input is the node representation produced by the preceding layer. For the first layer, the node representation is provided by a feature vector associated with each node. Some of the most commonly used GNN models include the Graph Isomorphism Network (GIN)~\cite{hamilton2017inductive}, Graph Attention Network (GAT)~\cite{velivckovic2017graph}, and Graph Convolutional Network (GCN)~\cite{kipf2016semi}.

The processing of each layer, as depicted in Figure~\ref{fig:gnn}, is divided into two phases. The first phase, known as the aggregation phase, involves a Sparse-Dense Matrix Multiplication (SpMM) between the graph adjacency matrix $A$, where $a_{i,j} = 1$ if nodes $i$ and $j$ are connected, and the set of node features from the previous iteration, $H^{(k-1)}$. The graph adjacency matrix is typically stored in Compressed Row Storage (CRS) format to optimize space and computation efficiency \cite{garg2022understanding, yan2020hygcn, geng2020awb}. The subsequent phase, referred to as the combination phase, consists of a Dense-Dense Matrix Multiplication (GEMM) between the intermediate matrix and a weight matrix $W$. While matrix multiplication is one common method for processing GNNs, another approach is message passing, where nodes communicate with their neighbors to update their representations.

The key dimensions in the computations are: the number of nodes $V$, the number of node input features $F$, the number of node output features $G$, and the degree (number of neighbors) of each node $N_v$.

\subsection{Spatial Accelerators and GNN Dataflows}
\ab{a lot of explanation, reduce this in the same way using the figure as support}
\label{sec:acc}
Graph data is irregular when compared to image, audio and tabular sources. This motivated the rise of specific GNN accelerators \cite{yan2020hygcn,yan2020hygcn,sarkar2023flowgnn} featuring either fixed dataflows or adaptive capabilities, which outperform conventional GPU and CPU execution enabling high performance thanks to parallelization and energy efficiency due to high specialization. 

Differently from traditional DNN workloads such as fully connected and convolutional layers \cite{sze2017efficient} \pol{TODO: Missing citation}, a layer of a GNN is organized in two phases, and an \textit{intra-phase dataflow} must be selected for each of them. Figure~\ref{fig:dataflow}a shows the loop nest representation of the main computation of a GNN layer. Aggregation and combination are implemented as two loop nests executed sequentially. The matrices involved are the same as in Figure~\ref{fig:gnn}. In each phase, the intra-phase dataflow is specified by the loop ordering and the spatial loop tiling and unrolling. For the latter, the \texttt{parallel-for}s describe the parallel execution of a loop on several Processing Elements (PEs) by unrolling a specific dimension of the involved matrices. In the aggregation phase, the tile sizes are \texttt{T\_Va}, \texttt{T\_Fa} and \texttt{T\_N}, representing the tile sizes for the $V$, $F$ and $N$ dimensions, respectively. The number of maximum utilized PEs is $\texttt{T\_Va} \times \texttt{T\_Fa} \times \texttt{T\_N}$. However, notice that the $N_v$ dimension is the number of neighbors and depends on $v \in \mathcal{V}$, hence while iterating on low-degree nodes, the spatial accelerator may be underutilized with high  \texttt{T\_N}s. In the combination phase, the loop nest represents a dense matrix multiplication, and the tiling factors are \texttt{T\_Vc}, \texttt{T\_Gc} and \texttt{T\_Fc}, representing the tile sizes for the $V$, $G$ and $F$ dimensions respectively. In this phase, the static number of utilized PEs is equal to $\texttt{T\_Vc} \times \texttt{T\_Gc} \times \texttt{T\_Fc}$.

The global buffer and the PEs are interconnected through a Network-on-Chip (NoC) for data movement. According to the unrolling dimensions, data movement between PEs may be needed for spatial reduction, \ie accumulation of partial sums, or multicast capabilities could be useful to reduce memory accesses. Usually, each dataflow requires specific microarchitectural implementation.

Furthermore, the dataflows of the two phases are interdependent. The interaction between the aggregation and combination phases is described by the \textit{inter-phase dataflow} and determines the number of memory accesses needed to move data from one phase to the other. In this work, we adopt the sparse-dense workloads dataflow taxonomy proposed by Garg et al. \cite{garg2022understanding}. Figure~\ref{fig:dataflow} shows the space-time diagram of three different inter-phase dataflows:
 (i) \textbf{Sequential}~(\texttt{seq}). With this configuration, the two phases are run sequentially similarly to two DNN layers. The output of the first phase (intermediate matrix) is written to memory and then loaded back to PEs for the next phase. The size of the intermediate matrix is $V \times F$; hence, as shown in Figure~\ref{fig:dataflow}b, for large graphs, the intermediate matrix cannot be stored entirely in the global buffer, requiring additional accesses to the off-chip DRAM causing higher energy consumption.
(ii) \textbf{Sequential Pipeline}~(\texttt{sp}). Applying loop fusion and temporal tiling techniques to the Sequential loop nest, the execution of the Aggregation and Combination can be split in smaller steps temporally interleaved. As shown in Figure~\ref{fig:dataflow}c, at each step, a small tile of the intermediate matrix is calculated and stored in the global buffer or even in PE local buffers  
to be used in the next phase step without accessing higher level memory.
(iii) \textbf{Parellel Pipeline}~(\texttt{pp}). As shown in Figure~\ref{fig:dataflow}d, the two phases can also be executed in parallel. This requires partitioning PEs between two phases, the global buffer working as a ping-pong buffer and the NoC supporting the dual data movement. Balancing the two pipeline stages is crucial to avoid stall cycles. In particular, the intermediate matrix tile production rate of the aggregation phase should be equal to the combination consumption rate. Achieving this can be challenging, especially with heterogeneous node degree distributions.

\begin{figure}[t]
    \centering
    \includegraphics[width=\linewidth]{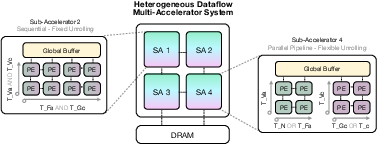}
    \vspace{-15pt}
    \caption{A multi-accelerator system for GNN inference.}
    \label{fig:hda}
    \vspace{-0.5cm}
\end{figure}

Similarly to the traditional DNN acceleration domain~\cite{garg2022understanding}, multiple accelerators, each featuring a diverse set of supported dataflows and flexibility degrees, can be combined together to compose a heterogeneous multi-accelerator system such as the one shown in Figure~\ref{fig:hda}. Such systems can be very effective in highly parallel use cases, allowing multi-dataflow execution without the hardware overhead needed in highly flexible single-accelerator systems.

\subsection{Motivation}\label{subsec:problem-statement}

Despite advancements in accelerator design for GNNs, the main challenge in both flexible dataflow single-accelerator and multi-accelerator heterogeneous dataflow systems for GNNs remains the following: differently from DNN workloads, featuring fixed tensor sizes independent of the input, in GNNs, the amount of computation involved and consequently the size and structure of the matrices involved varies on the specific input graph (especially $V$ and $N_v$s). Variations in graph structures can cause substantial performance differences across dataflow strategies used by accelerators.

\begin{figure}[t]
    \centering
    \includegraphics[width=0.9\linewidth]{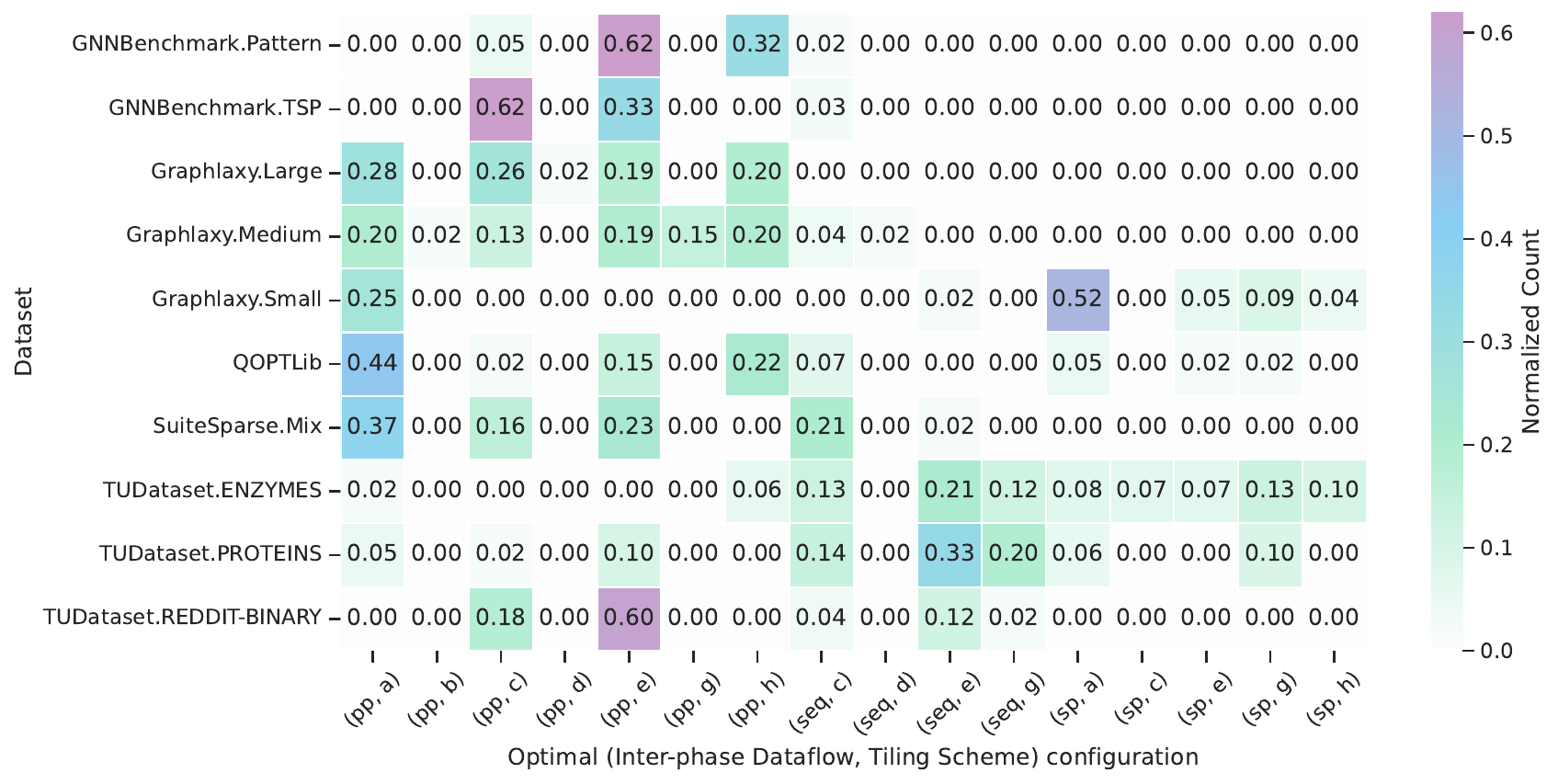}
    \vspace{-10pt}
    \caption{Frequency of optimal dataflow configurations.}
    \label{fig:best_configuration}
    \vspace{-0.5cm}
\end{figure}

Thus, while for DNN workloads, an optimal dataflow choice for each layer can be determined offline and adopted in deployment for all the situational inputs \cite{mei2021zigzag, jung2023salsa} \pol{TODO: cite this}, for GNNs there is no one-fits-all solution. The optimal configuration of intra-phase dataflow and inter-phase dataflow has to be determined dynamically for each input dataset as it depends on the dataset characteristics. As shown in Figure~\ref{fig:best_configuration}, even for graphs of the same dataset, the optimal configuration in terms of latency varies, and different datasets exhibit different overall optimal configuration distributions.

The problem we aim to address in this work is predicting the latency of computing a GNN inference on a specific accelerator configuration. Having this knowledge prior to execution enables better computational decisions around GNN workloads. We are especially interested in using those predictions to rank configurations to select which is the best configuration for a given GNN.

For a more specific application, we consider the scenario of scheduling a set of jobs, namely GNN layers, each having a release time $r_j$ (arrival time) not known in advance, in order to minimize the average completion times ~\cite{phillips1998minimizing}. The objective is to address the real-time scheduling problem for heterogeneous multi-accelerator setups. This real-time scheduling scenario has been extensively investigated for traditional DNN workloads, with particular emphasis on Inference-as-a-Service and Multi-Tenant Data Center applications~\cite{kim2023moca, blanco2024online}. 
The problem can be considered as unrelated parallel machine scheduling with job release times, which is NP-hard even in an offline setting and with arrival times known a priori~\cite{schulz2002scheduling, lenstra1977complexity}. Leveraging the proposed latency predictor, machine-learning based speed-oblivious online scheduling techniques can be adopted~\cite{lindermayr2023speed}.

\section{Data-driven GNN Dataflow Selection}
We address the challenge of predicting the latency of executing a GNN with a specific accelerator configuration (Section \ref{subsec:problem-statement}) by modeling dataflow selection as a learning problem. For an intra-dataflow $\omega \in \Omega$ and an inter-phase dataflow $\delta \in \Delta$ our objective is to select the set of parameters $\theta$ attaining the best approximation for the latency $l$ of execution
\begin{equation} \label{eq:latency-estimation}
    \pi_\theta(\mathcal{G}, \omega, \delta) \simeq l(\mathcal{G}, \omega, \delta).
\end{equation}
Knowing the latency before execution enables making dataflow-aware decisions for a collection of graphs. In this section, we first explore how to successfully find $\theta$ using supervised learning, and we then provide experimental results supporting our claims.

\subsection{Learning to Predict Latency}
\label{subsec:latency-prediction}

\begin{table}[t]
    \caption{Proposed latency prediction model input features.}
    \label{tab:features}
    \centering
    \resizebox{0.9\columnwidth}{!}{
    \begin{tabular}{@{}l m{0.7\columnwidth} c c@{}}
        \toprule
         & Description & \multicolumn{2}{c}{Symbol and/or Equation} \\
         \midrule
         \midrule
         \parbox[t]{2mm}{\multirow{5}[20]{*}{\rotatebox[origin=c]{90}{base}}} & Number of nodes and edges & \multicolumn{2}{c}{$V$ and $E$} \\
         \cmidrule{2-4}
         & Spatial tiling factors & \multicolumn{2}{c}{$\texttt{T\_Va}, \texttt{T\_Fa}, \texttt{T\_N}, \texttt{T\_Vc}, \texttt{T\_Gc}, \texttt{T\_Fc}$}  \\
         \cmidrule{2-4} %
         & Graph density & \multicolumn{2}{c}{$E \times V^{-2}$} \\
         \cmidrule{2-4} %
         & Clustering coefficient, measuring the probability that the adjacent nodes of a node are connected & \multicolumn{2}{c}{\cite{wassington2022bias}} \\
         \cmidrule{2-4} %
         & Seven node degrees quantiles normalized with respect to maximum degree including minimum degree & \multicolumn{2}{c}{$Q_i, \quad \forall i \in \{1, \ldots, 7 \}$} \\
         \midrule \midrule
          \parbox[t]{2mm}{\multirow{5}[30]{*}{\rotatebox[origin=c]{90}{base+features}}} & Number of operations, assuming a dense matrix mul. for the aggregation phase & $S_1$ & $V \times F \times \left( G  + \text{Mean Degree}\right)$ \\
          \cmidrule{2-4} %
          & Estimation of the number of cycles for the combination phase & $S_2$ & $\frac{V \times F \times G}{\texttt{T\_Vc} \times \texttt{T\_Fc} \times \texttt{T\_G} } $ \\
          \cmidrule{2-4} %
          & Estimation of the number of cycles for the aggregation phase (dense mat. mul.) & $S_3$ & $\frac{V \times F \times \text{Mean Degree}}{\texttt{T\_N} \times \texttt{T\_Fa}}$ \\
          \cmidrule{2-4} %
          & Estimation of the latency for sequential inter-phase dataflow & $S_4$ & $S_1 + S_3$ \\
          \cmidrule{2-4} %
          & Cycles estimation for aggregation phase assuming CSR encoding & $S_5$ & $\sum_v^\mathcal{V} \frac{N_v \times F}{\min \left( N_v, \texttt{T\_N} \right) \times \texttt{T\_F}}$ \\
          \cmidrule{2-4} %
          & Cycles estimation for sequential inter-phase dataflow assuming CSR encoding & $S_6$ & $S_3 + \frac{S_5}{\texttt{T\_Va}}$ \\
    
         \bottomrule
    \end{tabular}
    }
    \vspace{-0.3cm}
\end{table}

In order to learn the GNN and configuration to latency mapping $\pi_\theta : \mathcal{G} \times \Omega \times \Delta \rightarrow \mathbb{R}$ effectively for any dataset, we require: (i) a significant amount of data with variability in the feature space of the graphs $\mathcal{G}$ topology, and (ii) the latency values for such GNN data across all accelerator configurations in $\Omega \times \Delta$. We satisfy the requirements by combining two tools. For the graph level variability, we use Graphlaxy \cite{wassington2022bias}, a graph dataset generator where the resulting datasets are spread in the feature space of the graphs. We obtain the latencies of executing a graph across different accelerator configurations with STONNE-Omega \cite{garg2022understanding}, a framework for accurate simulation of the latency $\hat{l}(\mathcal{G}, \omega, \delta)$ of executing a graph $\mathcal{G}$ with an intra-dataflow $\omega$ and an inter-phase dataflow $\delta$. By simulating the Graphlaxy graphs on STONNE-Omega we can generate a comprehensive dataset with variability in both $\mathcal{G}$ and $\omega, \delta$ axis. Across the dataset, we select the parameters $\theta$ that satisfy
\[
\argmin_{\theta}[\pi_\theta(\mathcal{G}, \omega, \delta) - \hat{l}(\mathcal{G}, \omega, \delta)]^2.
\]

Our model $\pi_\theta$ begins with a graph encoding step where metrics of interest are extracted from the graph $\mathcal{G}$. These metrics include conventional graph measures such as the number of nodes, density, and clustering coefficient, along with custom metrics engineered with the learning goal in mind (see Table~\ref{tab:features} for a complete list of definitions). The encoding step is followed by a learnable non-linear transformation. We find gradient boosting methods to be the best-suited models as they provide accurate results with very fast training and inference. These methods proved particularly effective in handling the large variability of latency magnitude, outperforming other approaches, such as MultiLayer Perceptrons (MLPs), which required accurate data normalization to achieve comparable results. Additionally, we find that using separate models for each configuration results in more accurate predictions. By using this approach, we can effectively predict the latency of GNN executions across various accelerator configurations, allowing for optimized scheduling and resource allocation in heterogeneous computing environments.

\subsection{Experiments on Latency Prediction} \label{subsec:experiments_latencyprediction}

\begin{table}[t]
    \centering
    \caption{The 8 tiling schemes considered in this work.}
    \label{tab:tilings}
    \resizebox{0.9\columnwidth}{!}{
    \begin{tabular}{l c c c c c c}
        \toprule
        Tiling & \multicolumn{3}{c}{Aggregation} & \multicolumn{3}{c}{Combination} \\
        \cmidrule(lr){2-4} 
        \cmidrule(lr){5-7} 
        
         $\omega$ & $\texttt{T\_Va}$ & $\texttt{T\_Fa}$ & $\texttt{T\_N}$ & $\texttt{T\_Vc}$ & $\texttt{T\_Gc}$ & $\texttt{T\_Fc}$ \\
         \midrule
         a & $*$ & $ \min \left(\text{Num. PEs}, F\right) $  & $1$ & $*$ & $1$ & $ \min \left(\text{Num. PEs}, F\right) $ \\
         b & $*$ & $ \min \left(2, F\right) $  &  $\lfloor F / 2 \rfloor$ & $*$ & $1$ & $ \min \left(2, F\right) $ \\
        
         c & $*$ & $ \min \left(8, F\right) $  &  $\lfloor F / 2 \rfloor$ & $*$ & $1$ & $ \min \left(8, F\right) $ \\

         d & $*$ & $1$  &  $1$ & $*$ & $1$ & $1$ \\

         e & $*$ & $ \min \left(18, F\right) $  &  $\lfloor F / 2 \rfloor$ & $*$ & $1$ & $ \min \left(18, F\right) $ \\
         f & $*$ & $ 1 $  &  $\min \left( 18, V \right) $ & $*$ & $1$ & $1$ \\

         g & $*$ & $ \min \left(18, F\right) $  &  $1$ & $*$ & $1$ & $\min \left(85, F\right)$ \\

         h & $*$ & $1$  &  $\min \left( 18, V \right) $ & $*$ & $1$ & $\min \left(85, F\right)$ \\
         \bottomrule
    \end{tabular}
    }
    \vspace{-0.3cm}
\end{table}

In the following paragraphs, we present a comprehensive evaluation of our approach to predicting GNN execution latency using datasets generated by Graphlaxy for training. 

We experiment on 24 different configurations of STONNE-Omega given by the combination of the 3 inter-phase dataflows introduced in Sec.~\ref{sec:acc} and 8 tiling schemes. These schemes are reported in Table~\ref{tab:tilings} and were selected based on insights from~\cite{garg2022understanding}, considering different granularities and both spatial and temporal aggregation of neighbours. For all the tilings, the $*$ symbol indicates that PE utilization is maximized using \texttt{T\_Va} and \texttt{T\_Vc}. In our STONNE-Omega simulations, we employ the same accelerator architecture as in~\cite{garg2022understanding}, comprising 512 PEs, each equipped with a 64B local register file.

We consider GCN~\cite{kipf2016semi} as the GNN model for our experiments. We use lightGBM~\cite{ke2017lightgbm} for the gradient boosting method implementation.

We propose two objectives for this area of experimentation. The first objective is to evaluate how well models trained on labels from executions of Graphlaxy graphs predict the latencies of unseen graphs coming from both Graphlaxy and other distributions, including graphs from real-world datasets. We evaluate the quality of the regressions by measuring the MAPE. The lower the MAPE, the tighter the predictions obtained by the model. The second objective is to assess the impact of using the trained models to predict the best configuration for a single accelerator. In order to measure how good are the best configurations suggested by the model we measure the top-k accuracy of the predicted best configuration being on the top-k of true best accelerators. For deeper insight into what improvement would this selections achieve we compare the latencies of a flexible accelerator guided by our model against three benchmarks for alternative dataflow selection strategies: (i) random, selecting a random configuration for each graph; (ii) best fixed, selecting only one configuration which works best across the specific dataset; (iii)~optimal, selecting the true best configuration for each graph.

We use three Graphlaxy datasets simulated on 24 STONNE-Omega configurations: small ($243$ graphs, avg. $10.49$ nodes), medium ($9,171$ graphs, avg. $241.95$ nodes), and large ($6,381$ graphs, avg. $782.36$ nodes). We also include datasets derived from various real-world distributions, including 43 graphs representing real-world networks sourced from the SuiteSparse Matrix Collection~\cite{suitesparse}, and 40 graphs from the QOPTLib~\cite{osaba2023qoptlib} dataset, which encompasses quantum combinatorial optimization problem representations. Additionally, we consider 600 enzyme representations, 1113 protein representations, and a collection of 2000 graphs representing Reddit discussions from~\cite{tudataset}. Furthermore, we include the Pattern and Travelling Salesman Problem (TSP) semi-synthetic datasets as described in~\cite{dwivedi2022benchmarkinggraphneuralnetworks}. The training and validation are done exclusively on the Graphlaxy graphs with the objective of obtaining models with low graph structure bias. We utilize a (70, 15, 15) split for training, validation and testing on the Graphlaxy datasets. All other non-Graphlaxy datasets are considered exclusively for testing.

Table \ref{tab:omega_latency_prediction_results} shows the experimental results in latency prediction metrics for the test sets. For regression quality, we observe a very small MAPE under $5\%$ in medium and large in-distribution graphs, the error increases for out-of-distribution graphs and also for smaller graphs independently. The models are able to predict the best configuration out of the 24 possible configurations at different rates across different datasets. There are significant improvements from predicting the best configuration to predicting a Top-3 configuration, the magnitude of this improvements varies across datasets. We find that a smaller MAPE does not always lead to better Top-k accuracy; for instance, the TSP dataset has a larger MAPE than Enzimes but more than double the Top-k accuracy. The improvement of a model-guided flexible accelerator against random choices is large and consistent, over $80\%$ across all datasets. Against the best fixed configuration, the model-guided flexible accelerator yields improvements for 8 out of 10 datasets. Naturally, fixed configurations show good performance for datasets with more homogeneous graphs, such as Enzymes and Pattern. Improvements over the best fixed configuration are also smaller for datasets with smaller graphs. Our method performs $5\%$ or less from optimal in half of the datasets and under $15\%$ from optimal in all datasets. The consistency in improvement over random and degradation over best is an indicator that, even when the models do not predict a Top-1 or Top-3 configuration all the time, there is a small performance difference between the predicted configuration and the best selection and a significant difference between the predicted configuration and a random choice.

\begin{table}[t]
    \caption{Proposed methodology in predicting GCN layer latency and identifying optimal dataflow configurations.}
    \label{tab:omega_latency_prediction_results}
    \centering
    \resizebox{\columnwidth}{!}{
        \begin{tabular}{l |c |c |c |c |c |c}
            \toprule
                \multirow{2}{*}{Dataset Name} & \multirow{2}{*}{MAPE $\downarrow$} & Top-1 & Top-3 &  \multicolumn{2}{c|}{Improvement over} & Degradation \\
                        & & acc. (\%) $\uparrow$ & acc. (\%) $\uparrow$ & random (\%) $\uparrow$ & best fixed (\%) $\uparrow$ & over optimal (\%) $\downarrow$ \\
            \midrule
            Graphlaxy.Medium & 3.78 & 91.28 & 99.62 & 93.63 & 58.42 & 0.56 \\
            Graphlaxy.Large & 4.83 & 83.63 & 98.20 & 96.27 & 8.99 & 0.86 \\
            Graphlaxy.Small & 17.36 & 46.94 & 75.51 & 97.93 & 0.49 & 5.12 \\
            SparseSuite.Mix & 13.32 & 62.79 & 81.40 & 89.79 & 8.04 & 5.24 \\
            QOPTLib & 24.99 & 56.41 & 74.36 & 82.45 & 20.98 & 14.94 \\
            PATTERN & 40.17 & 33.13 & 68.56 & 84.04 & -6.30 & 9.40 \\
            TSP & 29.13 & 62.18 & 92.85 & 87.38 & 2.18 & 4.22 \\
            ENZYMES & 20.23 & 19.52 & 40.41 & 91.95 & -3.29 & 12.10 \\
            PROTEINS & 15.82 & 30.45 & 56.76 & 91.49 & 1.24 & 11.56 \\
            REDDIT-BINARY & 14.16 & 35.08 & 76.45 & 85.96 & 0.74 & 8.68 \\
            \bottomrule
        \end{tabular}
    }
    
\end{table}

Table \ref{tab:omega_latency_prediction_results_all_models_small_version} shows ablation results in justification of our modeling decisions. We observe that the base model requires some iteration in order to achieve the reported performance. Both regression and selection capabilities improve when adding synthetic features and when applying a logarithm transformation.
\begin{table}[t]
    \caption{Ablation study on latency regression with added composite features (\textit{+features}) and log-transformed target (\textit{+log}).}
    \centering
    \resizebox{\columnwidth}{!}{
        \begin{tabular}{l |c |c |c }
            \toprule
            Dataset Name & Model Name & MAPE $\downarrow$ & Degradation over optimal (\%) $\downarrow$ \\
            \midrule
            \multirow{4}{*}{Graphlaxy.Medium} & base & 35.77 & 48.12 \\
             & base+features & 10.76 & 4.06 \\
             & base+log & 4.95 & 0.84 \\
             & base+features+log & 3.78 & 0.56 \\
            \midrule
            \multirow{4}{*}{SparseSuite.Mix} & base & 97.04 & 92.68 \\
             & base+features & 31.68 & 12.52 \\
             & base+log & 19.09 & 5.87 \\
             & base+features+log & 13.32 & 5.24 \\
            \midrule
            \multirow{4}{*}{QOPTLib} & base & 515.94 & 92.94 \\
             & base+features & 152.67 & 86.60 \\
             & base+log & 38.22 & 21.18 \\
             & base+features+log & 24.99 & 14.94 \\
            \bottomrule
        \end{tabular}
    }
    \label{tab:omega_latency_prediction_results_all_models_small_version}
    \vspace{-0.3cm}
\end{table}

\section{Online Scheduling}
As introduced in the problem statement (Section \ref{subsec:problem-statement}), we present a specific real-world application for our methodology of learning to predict latency. Consider an online scheduling scenario where we want to schedule a set of jobs with unknown arrival times under a Pareto distribution into heterogeneous accelerators to minimize the average completion times ~\cite{phillips1998minimizing}. In this section, we incrementally introduce an online scheduling strategy to leverage the latency prediction models from Section \ref{subsec:latency-prediction} along with a detailed experimental comparison against a selection of benchmarks.

\subsection{Latency Prediction Guided Online Scheduling}
\label{subsec:latency-prediction-guided-online-scheduling}
In the context of online scheduling, the Shortest Job First (SJF) heuristic is particularly effective because it minimizes the average waiting time for jobs. By prioritizing shorter jobs, SJF ensures that more tasks are completed in a given time frame, thereby reducing the overall time jobs spend in the queue. This approach is beneficial in environments with heterogeneous accelerators where job duration can vary significantly. The rationale behind SJF's effectiveness lies in its ability to keep the system's load balanced and prevent long jobs from delaying the completion of shorter ones. SJF can significantly outperform other scheduling algorithms in terms of minimizing average completion time, especially under high system loads and diverse job sizes \cite{schrage1968optimality}.

In most real scenarios we do not know the length of the job before execution. In practise, we use estimates of the job length to construct approximate SJF variants. A simple approach in a graph context is to use the number of edges or nodes to determine the length of a job. Such information on the graph size is useful for load balancing between accelerators. Nevertheless, the previous approach has no information about the specific graph-accelerator performance. We can overcome this limitation by using the latency prediction models as estimators for the job length. This approach provides a more accurate dataflow-aware estimation of the job length.

\subsection{Experiments on Online Scheduling}
\label{subsec:exp-onlinescheduling}
We assess the performance of online scheduling algorithms by measuring three metrics on jobs: 
(1) Mean Completion Time, when a job finishes;
(2) Mean Turnaround Time, difference between finishing time and arrival time; 
(3) Mean Execution Time, difference between finishing time and starting time.
We compare the following strategies to make scheduling decisions: 
(i) \texttt{Random};
(ii) \texttt{FCFS}, first come first serve; 
(iii) \texttt{LIFO}, last come first serve; 
(iv) \texttt{SJF-Nodes}, least nodes next; 
(v) \texttt{SJF-Edges}, least edges next; 
(vi) \texttt{SJF-Truth}, ground truth shortest job next; 
(vii) \texttt{\textbf{Ours} or \texttt{SJF-Predicted}}, the shortest job next based on latency predicted by the proposed model.
Note that \texttt{SJF-Truth} is the equivalent of the SJF heuristic introduced in Section \ref{subsec:latency-prediction-guided-online-scheduling} but is not feasible in practise since it requires knowledge about the true cost of executing a job in a processor. Our method serves as an approximation of SJF truth, which is feasible in practise.

We compose the online scheduling dataset for the experimentation. This dataset contains all the non-Graphlaxy graphs from the datasets from Section \ref{subsec:experiments_latencyprediction}. The inter-arrival times are randomly drawn from a Pareto distribution, yielding a mean utilization rate of $85\%$ across the included accelerator settings, and each graph corresponds to a job.

We consider a first scenario where we have a set of 3 accelerators (one for each inter-phase dataflow) and a flexible tiling configuration, similar to the one depicted in Figure~\ref{fig:hda}. Thus, in this scenario, we need to decide which graph to schedule in which accelerator and also which tiling configuration that accelerator will act upon. We schedule the graphs from the online scheduling dataset to the accelerators using all the previous algorithmic strategies separately. The accelerator tiling decision for each graph is decided using the prediction for our \texttt{SJF-Predicted} method, the optimal tiling for the \texttt{SJF-Truth} and a random tiling for other baselines. Note that the knowledge of the optimal tiling makes the algorithm theoretical; it is not feasible in practice; we mark such algorithms with \ding{70}.

\begin{figure}
    \centering
    \includegraphics[width=0.8\linewidth]{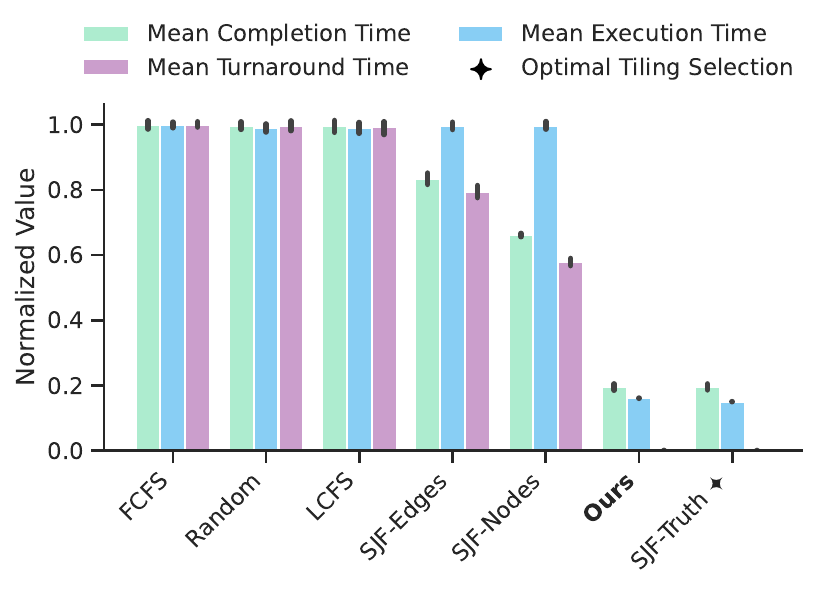}
    \vspace{-10pt}
    \caption{Algorithm performance for online scheduling policies with baselines using random tiling selection.}
\label{fig:online_scheduling_flexible_tiling_baselines_random}
\vspace{-0.3cm}
\end{figure}

Figure \ref{fig:online_scheduling_flexible_tiling_baselines_random} contains the online scheduling results for this first scenario over five runs with a fixed seed for all algorithms. Metrics are normalized separately by the largest value in any algorithm. We observe that the combined effects of both having better scheduling decisions and a dataflow-aware tiling selection yield a large improvement over baselines in all metrics while matching the performance of \texttt{SJF-Truth} first with optimal tiling selection.

Next, we are interested in understanding what magnitude of the improvements is obtained by the tiling selections versus having better estimates for the graph cost. We consider the same heterogeneous multi-accelerator setting with the same algorithmic baselines with the difference that now all baseline algorithms select the optimal tiling after assigning a graph to an accelerator. For our method the tiling is still selected with the prediction.

\begin{figure}
    \centering
    \includegraphics[width=0.8\linewidth]{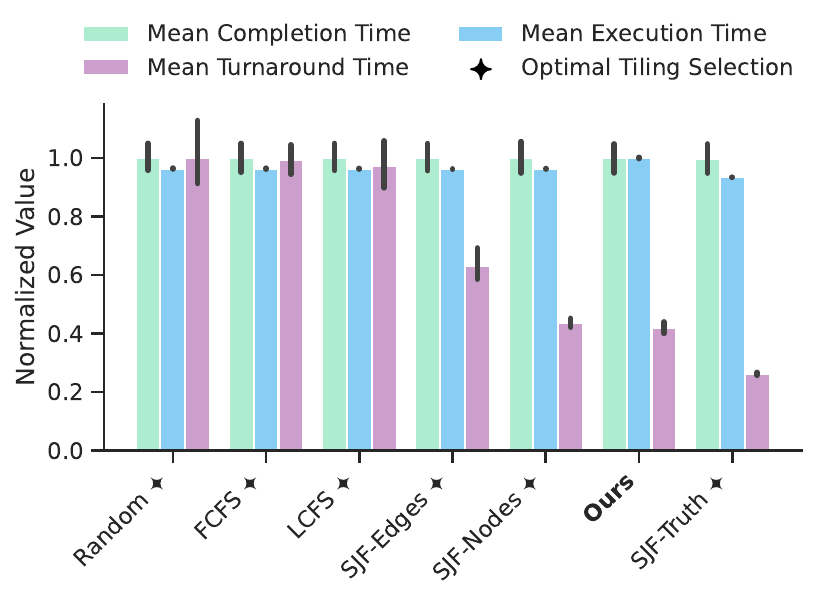}
     \vspace{-10pt}
    \caption{Algorithm performance for online scheduling policies with baselines using theoretical best tiling selection.}
\label{fig:online_scheduling_flexible_tiling_baselines_best}
\vspace{-0.7cm}
\end{figure}

Figure \ref{fig:online_scheduling_flexible_tiling_baselines_best} shows the results for the second setting. With optimal tiling decisions the mean execution times and mean completion times of all baselines are very close to the true shortest job first performance. We observe that algorithms with better estimates for the graph cost significantly improve the turnaround time, indicating a superior ability to handle job queues and, consequently, better overall scheduling capabilities. Note that all methods using the true optimal tiling are not feasible in practise and thus marked with \ding{70}.

\paragraph{Predictor Runtime Considerations}

The computational cost of inference for predictors is an important consideration for the practical feasibility of our approach. In our implementation, we utilize a lightGBM boosting trees compiler called \textit{lleaves} \cite{lleaves} to compile the trained models. To assess the inference time, we conducted measurements on a workstation equipped with a Ryzen 9 5900x processor. The average inference time across all graphs in our datasets was $2,311.8803$ nanoseconds. Assuming an accelerator operating at $1$~GHz, and all the 24 models running in parallel on multiple cores, this translates to approximately $2,312$ accelerator cycles per inference.
To contextualize this performance, we compare it to the average waiting time for job scheduling in our previous \texttt{SJF-Predicted} runs, which was $18,756.49$ cycles. The inference time for our predictive models is thus only about $12.3\%$ of the average job waiting time. This comparison demonstrates the feasibility of implementing \texttt{SJF-Predicted} scheduling, even without dedicated GPU acceleration for the predictive models.
It is worth noting that the computational cost of model inference remains constant regardless of graph size. Consequently, as graph sizes increase, the relative cost of model inference becomes increasingly negligible in the overall scheduling process.

\section{Conclusion}
GNNs present state of the art performance across domains with relational nature. Recent efforts have shown that different accelerators with GNN-specific dataflow architectures significantly outperform conventional GPU and CPU acceleration. In this work, we introduce a novel approach to tackling the unexplored question of how graphs with different properties perform on different dataflows. We train models to predict the latency of executing a certain graph on a certain dataflow on an extensive dataset of simulations. Our experimental evaluation shows that we can efficiently predict the best dataflow for a given graph. This expands the potential of adaptability of GNN accelerators to graph inputs. Moreover, we introduce an online scheduling algorithm leveraging the predictors. Experimental results show that our method outperforms all feasible baselines and matches the performance of the shortest job first, strong but unfeasible in practise, heuristic. Our method achieves up to $3.17\times$ speedup in mean completion time, $6.26\times$ speedup in mean execution time and more than $1000\times$ speedup in turnaround time against the best feasible baseline on the online scheduling dataset.

Several intriguing research directions emerge for future
work. The first is to expand on a larger design space exploration of the architecture. The second is to consider more complex models for latency prediction and online scheduling although this direction would require additional time cost concerns.

\begin{acks}
This work has been (partially) supported by the Spoke 1 "FutureHPC \& BigData" of the Italian Research Center on High-Performance Computing, Big Data and Quantum Computing (ICSC), as well as by the EU through the European Research Council (ERC), grant ERC-StG-2021-WINC-101042080.
\end{acks}
\bibliographystyle{ACM-Reference-Format}
\bibliography{bib}
\end{document}